\documentclass[letterpaper]{article} 
\usepackage{aaai25}  
\usepackage{times}  
\usepackage{helvet}  
\usepackage{courier}  
\usepackage[hyphens]{url}  
\usepackage{graphicx} 
\usepackage{natbib}  
\usepackage{caption} 
\usepackage{algorithm}
\usepackage{algorithmic}
\usepackage{amsmath}
\usepackage{amssymb}
\usepackage{multirow}
\usepackage{booktabs}
\usepackage{newfloat}
\usepackage{listings}
\DeclareCaptionStyle{ruled}{labelfont=normalfont,labelsep=colon,strut=off} 
\floatstyle{ruled}
\newfloat{listing}{tb}{lst}{}
\floatname{listing}{Listing}
\title{Falcon: Faster and Parallel Inference of Large Language Models Through Enhanced Semi-Autoregressive Drafting and Custom-Designed Decoding Tree}
\author{
   Xiangxiang Gao\equalcontrib,
   Weisheng Xie \equalcontrib\thanks{Corresponding Author}, 
   Yiwei Xiang, 
   Feng Ji
}
\affiliations{
  Bestpay AI Lab, Shanghai, 200080 China\\
\{gaoxiangxiang, xieweisheng, xiangyiwei, jifeng\}@bestpay.com
}

\usepackage{bibentry}

\begin{document}

\maketitle

\begin{abstract}
Striking an optimal balance between minimal drafting latency and high speculation accuracy to enhance the inference speed of Large Language Models remains a significant challenge in speculative decoding. In this paper, we introduce Falcon, an innovative semi-autoregressive speculative decoding framework fashioned to augment both the drafter's parallelism and output quality. Falcon incorporates the Coupled Sequential Glancing Distillation technique, which fortifies inter-token dependencies within the same block, leading to increased speculation accuracy. We offer a comprehensive theoretical analysis to illuminate the underlying mechanisms. Additionally, we introduce a Custom-Designed Decoding Tree, which permits the drafter to generate multiple tokens in a single forward pass and accommodates multiple forward passes as needed, thereby boosting the number of drafted tokens and significantly improving the overall acceptance rate. Comprehensive evaluations on benchmark datasets such as MT-Bench, HumanEval, and GSM8K demonstrate Falcon's superior acceleration capabilities. The framework achieves a lossless speedup ratio ranging from 2.91x to 3.51x when tested on the Vicuna and LLaMA2-Chat model series. These results outstrip existing speculative decoding methods for LLMs, including Eagle, Medusa, Lookahead, SPS, and PLD, while maintaining a compact drafter architecture equivalent to merely two Transformer layers. Our code is publicly available at 
https://github.com/Bestpay-inc/Falcon.
\end{abstract}



\section{Introduction}
Large Language Models (LLMs) have demonstrated exceptional performance across various benchmarks, reinforcing their practical significance. These models are primarily built on the Transformer architecture and utilize autoregressive (AR) decoding, effectively capturing complex dependencies and generating coherent sequences \cite{wan2024efficientlargelanguagemodels}. However, due to their considerable model sizes, LLMs also face significant computational overhead and latency bottlenecks during inference. For example, the inference speed of the GLM 10B model is a mere 101 tokens per second when operated on a single Nvidia A100 GPU \cite{du2022glmgenerallanguagemodel}.
This presents considerable challenges for the widespread deployment and application of LLMs \cite{zhu2024surveymodelcompressionlarge,xiao2023surveynonautoregressivegenerationneural, 
ghazvininejad-etal-2019-mask,xu-etal-2021-distilled,DBLP:conf/iclr/DingW0WTT21,guo2021selfdistillationmixuptrainingnonautoregressive,Wang_Tian_He_Qin_Zhai_Liu_2019,10.1609/aaai.v33i01.33013723}. 


Intriguingly, the inference speed of LLMs is considerably hampered by memory bandwidth, with the main latency bottleneck stemming from reading and writing model parameters to memory, rather than from arithmetic computations \cite{xia2024unlockingefficiencylargelanguage}. To address this issue, researchers have proposed speculative decoding to expedite the inference speed of LLMs without sacrificing accuracy \cite{zhang2024draftverifylossless,Guo_Tan_Xu_Qin_Chen_Liu_2020}. This strategy employs a drafter that efficiently generates k tokens as speculation of future decoding steps within the LLMs. Subsequently, the LLM verifies these tokens, with only successfully validated tokens being accepted as decoded tokens, thus maintaining the quality of the generated output \cite{xia2023speculative}. By directing computational resources towards validating pre-generated tokens, speculative decoding significantly diminishes the memory operations needed to access LLM parameters, consequently boosting overall inference efficiency \cite{xia2024unlockingefficiencylargelanguage}. 

While speculative decoding presents a potential solution to decrease the inference latency of LLMs, it also raises several important questions that need further exploration. Notably, existing speculative decoding methods primarily employ two drafting strategies: AR and semi-autoregressive (SAR) drafting. AR drafting generates tokens sequentially, each dependent on the preceding ones. This sequential dependency constrains the parallelism of the draft models, leading to substantial time overheads \cite{guo2019finetuning,NEURIPS2019_74563ba2,geng-etal-2021-learning,10.1145/1143844.1143891}. In contrast, SAR drafting generates multiple tokens simultaneously, enhancing the parallelization of the drafting process. However, a significant limitation of SAR drafting is its inability to fully capture the inter-dependencies between tokens within the same block, potentially resulting in a lower acceptance rate for the generated tokens \cite{bao-etal-2022-textit,ran-etal-2020-learning}. Consequently, balancing low drafting latency with high speculation accuracy poses a significant challenge in speculative decoding aimed at accelerating the inference speed of LLMs \cite{xia2024unlockingefficiencylargelanguage}. 

In this paper, we introduce Falcon, an advanced SAR speculative decoding framework engineered to boost both the drafter's parallelism and the output quality, thereby amplifying the inference efficiency of LLMs. Falcon integrates the Coupled Sequential Glancing Distillation method, which elevates the token acceptance rate of SAR drafting. We offer an in-depth theoretical explanation for this accuracy enhancement, ascribing it to the strengthened dependencies between tokens within the same block. Additionally, we have developed a specialized decoding tree to support SAR drafting, enabling the drafter to generate multiple tokens in a single forward pass and also accommodate multiple forward passes. This design leads to a higher acceptance rate of tokens by the LLMs, further hastening inference speed.

Our comprehensive evaluations of Falcon on various benchmarks showcase its outstanding acceleration capabilities. It achieves a lossless speedup ratio ranging from 2.91x to 3.51x on the Vicuna and LLaMA2 model series. This performance outshines top-tier speculative decoding methods that use either AR drafting (such as Eagle, SPS, and PLD) or SAR drafting (like Medusa and Lookahead). Additionally, unlike other methods that necessitate billions of parameters for drafters, Falcon can attain faster inference speeds using parameters equivalent to just two Transformer blocks. Therefore, Falcon is highly beneficial for applications requiring real-time responses while working under computationally constrained environments. Our key contributions to the community are summarized as:

\begin{itemize}
    \item We introduce Falcon, an advanced SAR speculative decoding framework that enhances both the parallelism and output quality of the drafter. Falcon achieves a better equilibrium between reduced inference latency and elevated speculation accuracy.
    \item We develop Coupled Sequential Glancing Distillation, which markedly improves the accuracy of the SAR drafter. Additionally, a theoretical explanation is provided to clarify how this method enhances output quality.
    \item We design a specialized decoding tree to support SAR drafting, enabling the drafter to generate multiple tokens in a single forward and allowing multiple forward passes, resulting in an improved token acceptance rate and inference speed of the LLMs.
    \item Falcon outperforms existing speculative decoding methodologies for LLMs, attaining a speedup ratio ranging from 2.91x to 3.51x. Remarkably, it accomplishes this while preserving a compact drafter size, which is comparable to merely two Transformer layers.

\end{itemize}

\section{Related Work}
\subsection{Autoregressive Drafting}
Speculative decoding with AR drafting is a straightforward approach that employs a small language model as the drafter to generate tokens sequentially, each conditioned on its predecessors. The generated tokens are then validated by the LLMs to ensure alignment \cite{leviathan2023fastinferencetransformersspeculative,spector2023acceleratingllminferencestaged,chen2024cascadespeculativedraftingfaster, zhao2024lookaheadinferenceaccelerationframework}. SPS \cite{chen2023acceleratinglargelanguagemodel} is the pioneering work in this field, which generates draft tokens by invoking a 4B parameter AR drafter and validating these drafts using 70B parameter LLMs. PLD \cite{saxena2023prompt} replaces the draft model with simple string matching in the prompt to generate candidate token sequences. BiLD \cite{NEURIPS2023_7b97adea} utilizes the T5-small model to generate tokens, which are then validated by a T5-XXL model. SpecInfer \cite{10.1145/3620666.3651335} accelerates LLM serving through a tree-based inference and verification mechanism, organizing the drafts into a token tree. The correctness of all candidate token sequences is verified against the LLM using a tree-based decoding mechanism.

While AR drafting considerably boosts inference speed, it also imposes extra GPU memory costs, particularly for draft models with billions of parameters. Moreover, it adds time overhead from the drafter that could potentially counterbalance the advantages of the improved inference speed. EAGLE \cite{li2024eaglespeculativesamplingrequires}, the current state-of-the-art speculative decoding method with AR drafting, tackles the issue of high GPU memory usage by incorporating a token sequence advanced by one time step. However, due to its inherently sequential nature, EAGLE is still constrained by time overhead, which hampers its potential to further speed up the inference of LLMs.
\begin{figure*}[ht]
    \centering
    \includegraphics[width=1.7\columnwidth]{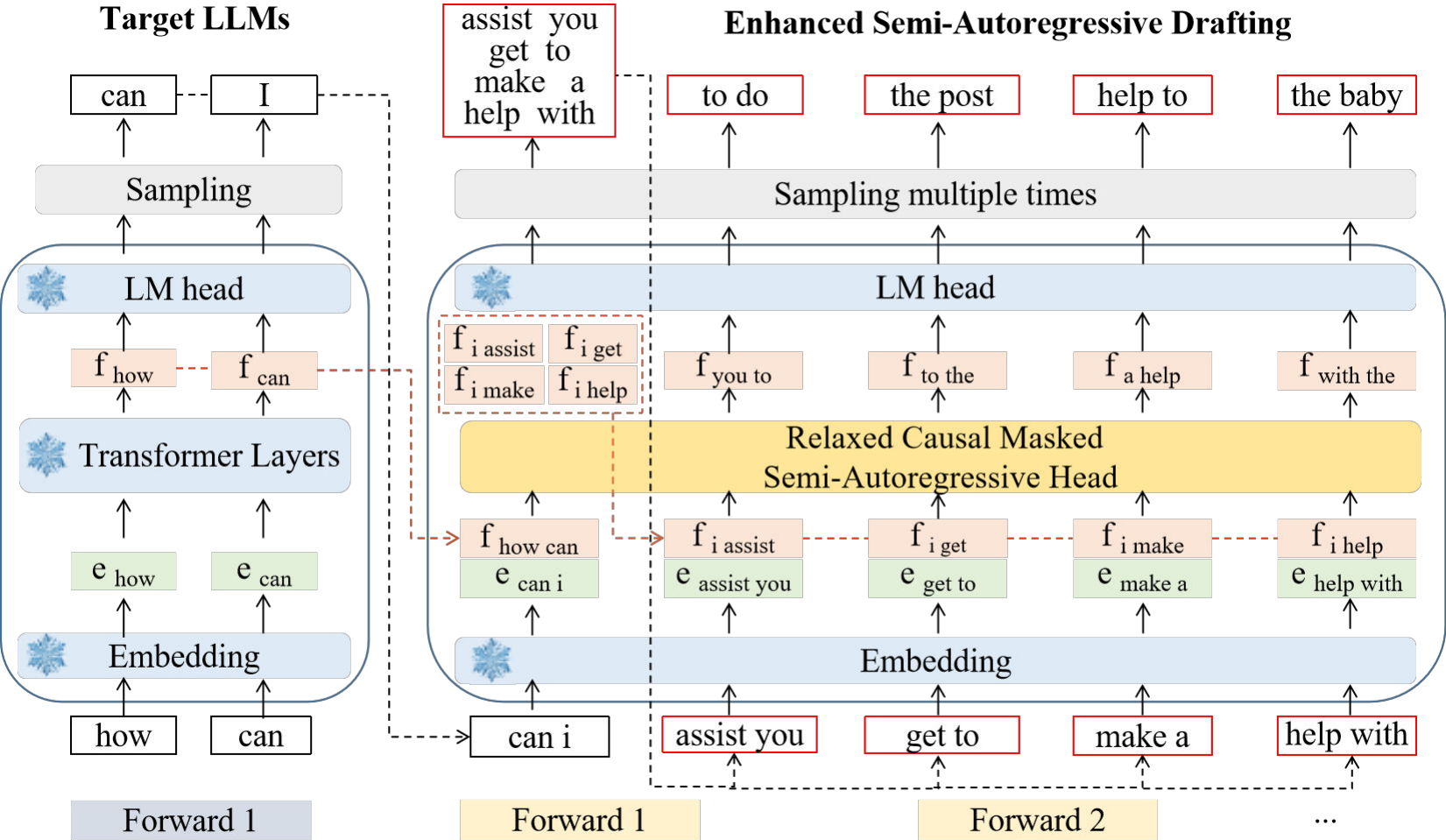}
    \caption{Framework of Falcon. It illustrates the computational process and displays the corresponding generation results of each forward pass for enhanced SAR drafting.}
    \label{fig:framework}
\end{figure*}

\subsection{Semi-autoregressive Drafting}
SAR speculative decoding simultaneously generates multiple tokens, maintaining the AR feature globally while easing it locally \cite{wang-etal-2018-semi-autoregressive,kaiser2018fastdecodingsequencemodels,pmlr-v80-oord18a,oord2016wavenetgenerativemodelraw}. Santilli et al. \cite{santilli-etal-2023-accelerating} propose that AR decoding can be restructured as parallel resolution of a non-linear equation through Jacobi and Gauss-Seidel fixed-point iterations. This technique directly appends multiple [PAD] tokens to the end of the input prompt, facilitating parallel generation and speeding up existing models. Lookahead \cite{zhao2024lookaheadinferenceaccelerationframework} utilizes this generation method to enable the LLMs to produce several separate n-grams concurrently within a single forward pass, thereby reducing the latency of LLM. PASS \cite{monea2023passparallelspeculativesampling} introduces multiple learnable tokens and fine-tunes these token embeddings to enhance parallel decoding performance. However, these methods deviate from the AR pre-training patterns, which can result in less optimal draft quality. 

Medusa \cite{cai2024medusasimplellminference} represents the most significant advancements in SAR drafting, building upon the research of Stern et al. \cite{NEURIPS2018_c4127b91}. It optimizes the process by freezing the backbone model and incorporating additional lightweight heads into it, enabling the concurrent generation of multiple tokens. Medusa effectively alleviates the computational cost typically associated with AR drafting, thus achieving remarkable speedups. However, its inference speed is constrained by the low accuracy of the drafter, which currently stands at 0.6. This drop in accuracy results from employing the parallel processing mechanism, whose predictions are solely based on input tokens without accounting for inter-token dependencies \cite{xia2023speculative,wertheimer2024acceleratingproductionllmscombined}. Therefore, the key to improving the output quality of SAR drafters lies in strengthening the inter-token dependencies within the same block. Such enhancements would enable an optimal balance between low drafting latency and high speculative accuracy.

\section{Falcon}
The Falcon framework is an advanced SAR speculative decoding architecture that concurrently improves the drafter's parallelism and token acceptance rate. A comprehensive overview and computational process of this framework are presented in Section \ref{chap:3.1}. To improve the drafter's accuracy, we introduce the Coupled Sequential Glancing Distillation (CSGD) method, which strengthens the inter-dependencies among tokens within each block. An in-depth description of the CSGD method is provided in Section \ref{chap:3.2}, followed by a theoretical discussion in Section \ref{chap:3.3}. In addition, a Custom-Designed Decoding Tree has been developed to support SAR drafting, which is described in Section \ref{chap:3.4}.

\subsection{Framework and Computational Process}\label{chap:3.1}

The architecture and computational process of the Falcon are depicted in Figure \ref{fig:framework}. It comprises three main components: the Embedding Layer, the Language Model (LM) Head, and the Relaxed Causal SAR Decoding Head. The Embedding Layer transforms a sequence of tokens into a corresponding sequence of token embeddings. The LM Head computes the probability distribution based on the aggregated features, from which the next token is sampled. Both the Embedding Layer and the LM Head leverage the parameters of LLMs. Relaxed Causal SAR Head will be introduced in section \ref{chap:3.2}.
\begin{figure*}[h]
    \centering
    \includegraphics[width=1.7\columnwidth]{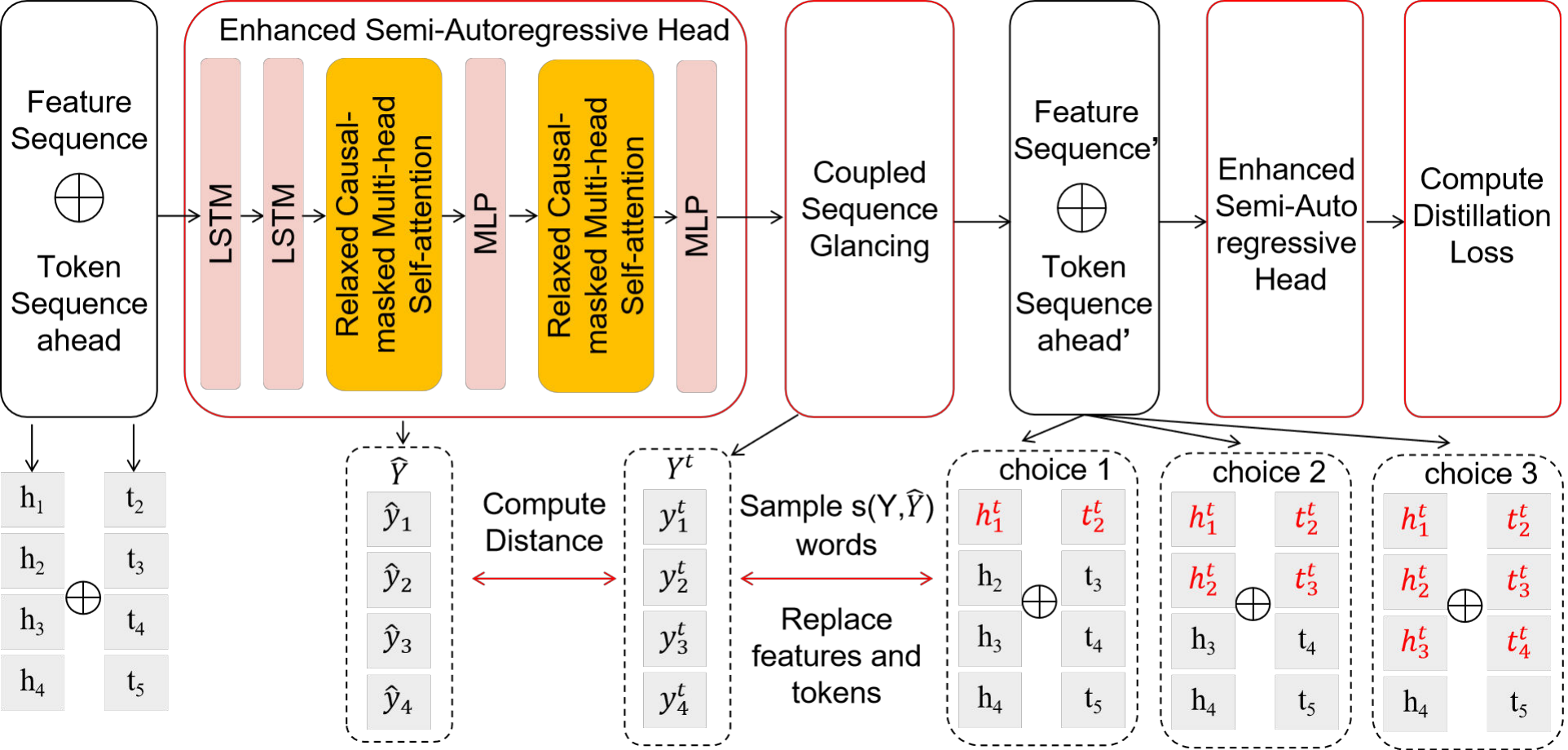}
    \caption{The training procedure of CSGD. $\hat{Y}$ is the initial predicted feature representation sequence of the draft model, $Y^{t}$ is the ground-truth feature calculated by LLMs, $t_i$ is the original token, $t_i^t$ is the target token generated by LLMs, $h_i$ is the original feature sequence, and $h_i^t$ is the target feature generated by LLMs.}
    \label{fig:training process}
\end{figure*}

EAGLE’s findings \cite{li2024eaglespeculativesamplingrequires} indicate that concatenating feature and token sequences from one time step ahead encapsulate richer semantic context. This enhancement allows the model to make more informed predictions by leveraging a broader scope of information. Consequently, we concatenate consecutive feature sequences and token sequences from one time step ahead to predict the next $k$ tokens concurrently. For instance, when $k=2$, Falcon predicts the feature sequence ($f_3, f_4$) using the initial feature sequence ($f_1, f_2$) and the token sequence ($t_2, t_3$) advanced by one time step. Subsequently, the predicted features ($f_3, f_4$), along with the next token sequence ($t_4, t_5$) are concatenated to form the new input sequence. This is used to predict subsequent feature sequences ($f_5, f_6$) and token sequences ($t_6, t_7$), facilitating the continuation of the drafting process.

\subsection{Coupled Sequential Glancing Distillation}\label{chap:3.2}
We have designed an SAR decoding method based on CSGD, aimed at enhancing the accuracy of the drafter. The training procedure is illustrated in Figure \ref{fig:training process}. Here, the feature and token sequences from one time step ahead are concatenated and fed into the drafter, resulting in a fused sequence of dimensions ($bs, seq\_len, 2*hidden\_dim$). The drafter is composed of a hybrid Transformer network, which includes two layers of LSTM \cite{hochreiter1997long}, Relaxed Causal-Masked Multi-Head Self-Attention, and MLP networks. The LSTM network reduces the dimension of the fused sequence to ($bs, seq\_len, hidden\_dim$) and retains information about past tokens, thereby improving the model's accuracy. The Relaxed Causal-Masked Multi-Head Self-Attention mechanism enables the model to focus on relevant parts of the input sequence while preserving causality. The MLP layers further process this information to make the final predictions. 

After the sequence passes through the drafter for the first time, an initial prediction of tokens, denoted as $\hat{Y}$ is generated. We compare the Hamming Distance \cite{roman1992coding} between the prediction from the drafter $\hat{Y}$ and the prediction from LLMs $Y^{t}$. Then, we replace a certain number of continuously predicted token sequence $t_i$ and features sequence $h_i$ with the correct ones $t_i^t$ and $h_i^t$ from LLMs. The number $N$ is computed as $N = \lambda\cdot d(Y^t, \hat{Y})$, where $d(\cdot)$ is the hamming distance, $\lambda=\frac{0.4*(ep_t-ep_c)}{ep_t}$, $ep_t$ is the total epoch number and $ep_c$ is the current epoch number. Note that our approach differs from the conventional glancing method \cite{qian-etal-2021-glancing}, which only replaces tokens randomly. Instead, we concurrently replace continuous token and feature sequences preceding those to be predicted, which is illustrated in the dashed boxes labeled Choice 1, 2, 3 in Figure \ref{fig:training process}. This enhances the comprehension of inter-token relationships, as well as ensures the drafter can effectively utilize token sequences from ahead time step. This is particularly beneficial in SAR decoding, where preserving the sequence’s integrity is crucial in maintaining optimum performance. Subsequently, the revised token and feature sequences are re-input into the drafter to compute the training loss. The training loss consists of two components: the regression loss and the distillation loss. For regression loss, we utilize the Smooth L1 loss:
\begin{align}
f_{  i:i+k}^{  draft}=Draft(t_{  i:i+k},f_{  i-1:i+k-1})\\
    L_{\text{reg}} = \text{Smooth L1}(f_{ i:i+k}, f_{  i:i+k}^{  draft})
\end{align}

Correspondingly, we optimize the drafter by calculating the distillation loss between the LLMs and the drafter:

\begin{align}
L_{\text{soft}}&=KL\_Div(p_{i:i+k},p_{  i:i+k}^{  draft}) \\
L_{\text{hard}}&=Cross\_Entropy(t_{i:i+k},t_{ i:i+k}^{ draft})\\
    L_{\text{dist}} &= \alpha L_{\text{soft}} + (1 - \alpha) L_{\text{hard}}
\end{align}

where $f$ and $f^{draft}$ denote features, $p$ and $p^{draft}$ denote the probability distribution, $t$ and $t^{draft}$ represent the tokens produced by the LLM and the drafter, respectively;  and $\alpha$, a constant coefficient set to 0.9. The losses $L_{soft}$ and $L_{hard}$, represent the soft and hard label losses as described in \cite{hinton2015distillingknowledgeneuralnetwork}, are independently computed using the Kullback-Leibler divergence and Cross-Entropy, respectively.

Using the same weight matrix of the LM Head, the logits can be calculated as follows:
\begin{align}
    p_{i:i+k}&=Softmax(LM(f_{i-1:i+k-1})) \\
p_{i:i+k}^{draft}&=Softmax(LM(f_{i-1:i+k-1}^{draft}))
\end{align}
By integrating regression and distillation loss, we train the SAR Head with the combined loss function:
\begin{align}
    L = L_{\text{reg}} + \omega_{\text{dist}} L_{\text{dist}}
\end{align}
$\omega_{dist}$ is set to 0.1. Moreover, we employ data augmentation by adding random noise sampled from a uniform distribution $U(-0.1,0.1)$ to avoid error accumulation in features.

\subsection{Theoretical analysis of CSGD}\label{chap:3.3}
We use a theory of information argument to illustrate the impact of CSGD. Take $k=2$ as an example, let $X$ represent the next token, $Y$ the second-next token, and $C$ the input context (Omitted from equations for simplicity). Traditional AR methods focus on $H(X)$, whereas SAR decoding with $k=2$ targets $H(X)+H(Y)$, decomposed as \cite{gloeckle2024betterfasterlarge,olsson2022incontextlearninginductionheads}:
\begin{align}
    H(X)=&H(X|Y)+I(X;Y)\\
    H(X) + H(Y) = & H(Y|X)+2I(X;Y)+H(X|Y)
\end{align}
In Equation (10), the left two terms represent the training loss of 2-token prediction models. The right terms decompose the loss into a local cross-entropy term for the prefix $(C,X)$, denoted as $H(Y|X)$, a mutual information term that captures the information about $Y$ contained in $X$, denoted as $I(X;Y)$, and term $H(X|Y)$ describes the uncertainty about $X$ given the prefix $C$ and suffix $Y$. We can observe that SAR decoding increases the weight of $I(X;Y)$. However, conventional training methods for SAR decoding typically consider only the classical next-token entropy $H(Y|X)$, while the terms $I(X;Y)$ and $H(X|Y)$ are always overlooked. As a result, these methods do not effectively learn the dependencies between tokens within a block. 
However, CSGD addresses this issue. When predicting $X$, CSGD can leverage the features and tokens from one time step ahead in $C$. The feature sequences represent the prefix, and the token sequences represent the mutual information capturing the correlation between $C$ and $X$. Therefore, the term $H(X|Y)$ should be changed to $H(X|C)$, and $I(X;Y)$ should be changed to $I(X;C)$. In addition, when predicting token $Y$, CSGD can see the information about $C$ and $X$ simultaneously. Therefore, the training loss of CSGD is changed to:
\begin{align}
    H(X) = &H(X|C) + I(X; C)\\
    H(X)+H(Y)=& H(X|C)+I(X; C)\notag\\
    &+H(Y|X)+I(X;Y)
\end{align}
Equations (11) and (12) denote CSGD improves the dependency between tokens within a block, making the training loss of the SAR approach more similar to that of the AR approach. As a result, the probability distribution of multiple tokens predicted simultaneously by SAR decoding becomes more aligned with the distribution of tokens predicted sequentially by AR decoding. It further increases the probability that the tokens generated by SAR drafting will be accepted by the LLMs.

\subsection{Custom-Designed Decoding Tree}\label{chap:3.4}
\begin{figure}[h]
    \centering
    \includegraphics[width=1\columnwidth]{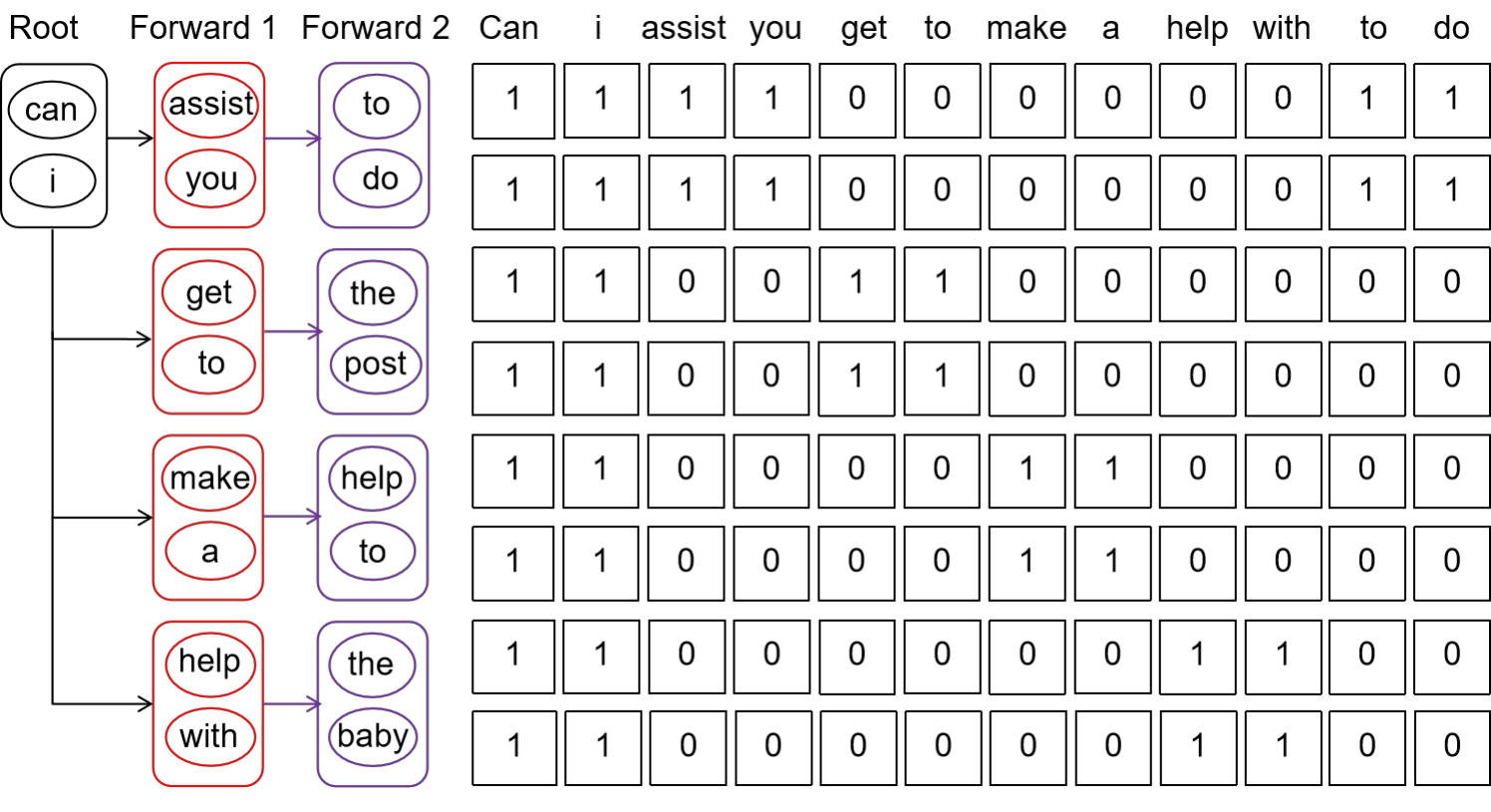}
    \caption{SAR decoding tree attention illustrated. This visualization demonstrates that SAR tree attention is utilized to process multiple candidates in parallel, and k is set to 2.}
    \label{decoding tree}
\end{figure}
\begin{figure*}[htbp!]
\centering
\includegraphics[width=1.9\columnwidth]{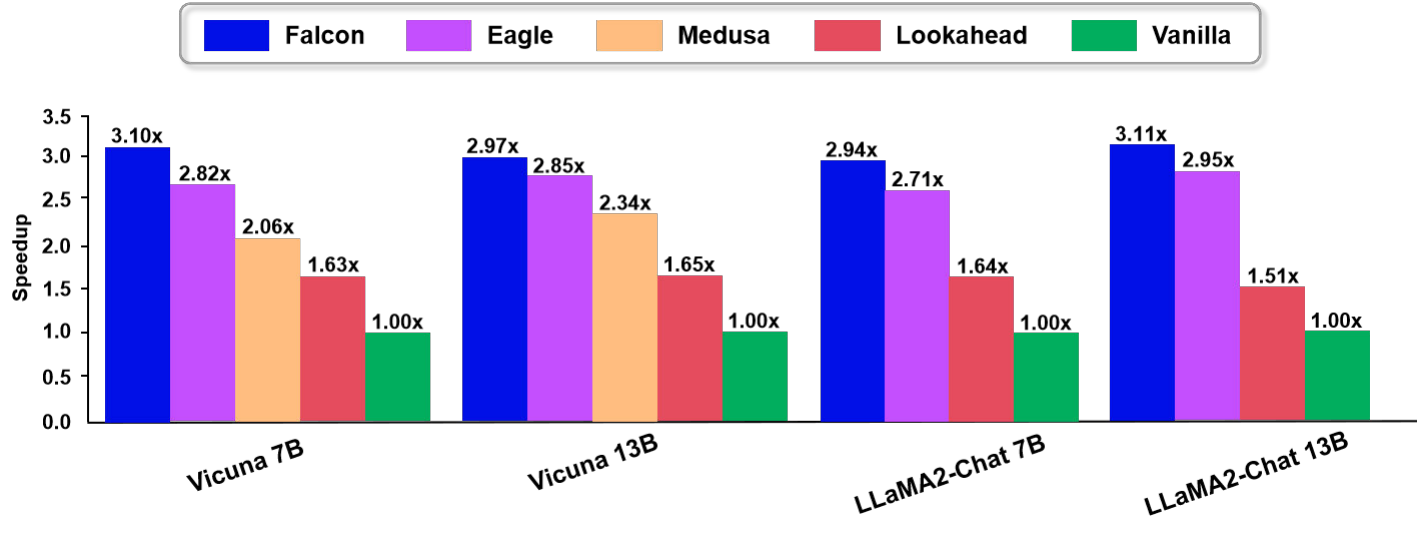} 
\caption{Speedup ratio of Vicuna and LLaMA2-Chat on MT-bench for greedy (temperature=0).}
\label{compare_figure}
\end{figure*}
A Custom-Designed Decoding Tree has been introduced to accommodate SAR drafting. It enables the drafter to perform m forward passes and generate n*k tokens for each forward pass, where n is the number of the tree nodes. These m*n*k tokens are then organized in a tree structure. Meanwhile, the generated tokens and the corresponding features are concatenated for the subsequent forward passes. In traditional AR decoding, a causal mask is used, structured as a lower triangular matrix. It ensures that the former tokens can not access the later information. In contrast, Falcon employs a Relaxed Causal Mask, which allows the model to access tokens within the same k*k block and their predecessors in the tree, illustrated in Figure \ref{decoding tree}.

Subsequently, the drafts are organized as a token tree, whose nodes each represent a candidate token sequence. The correctness of the candidate token sequences are verified by the LLMs using a tree-based parallel decoding mechanism, which is consistent with SpecInfer \cite{10.1145/3620666.3651335}. We record the accepted tokens and their corresponding features for the subsequent drafting phase. 

Our custom-designed decoding tree, engineered specifically for SAR, is more efficient than those used for fully AR approaches. In SAR decoding, a tree-structure drafter generates k times as many tokens as an AR decoding drafter in a single forward pass. As a result, m forward passes can propose k*m*n tokens, which is k times greater than an AR tree would generate. This enhancement significantly boosts the efficiency of token generation for drafters, allowing the LLMs to verify more tokens concurrently. Therefore, this improvement increases the likelihood of drafting tokens being accepted by the LLMs, thereby accelerating the overall inference speed of the LLMs.

\section{Experiments and Analysis}
\subsection{Models and Tasks}
We conducted experiments on Vicuna models (7B, 13B) and LLaMA2-Chat models (7B, 13B). We evaluated Falcon across multiple tasks, including multi-turn dialogue, code generation, and mathematical reasoning, employing the MT-bench \cite{zheng2024judgingMtBench}, HumanEval \cite{chen2021evaluatingHumanEval}, and GSM8K \cite{cobbe2021trainingGSM8k} datasets, respectively. We conducted experiments with a batch size of 1, a temperature of 0 (greedy decoding), and adopted experimental settings consistent with other works like Eagle and Medusa.
\subsection{Metrics}
Like other speculative decoding methods, Falcon primarily focuses on latency rather than throughput. We assess acceleration effects using the following metrics:
\begin{itemize}
\item \textbf{Wall-time speedup ratio}: The actual test speedup ratio relative to vanilla AR decoding.
\item \textbf{Acceptance rate ($\alpha$)}: The ratio of tokens generated by the drafter to all tokens gauges draft accuracy.
\item \textbf{Average acceptance length ($\tau$)}: The number of accepted tokens in each draft phase.
\end{itemize}

\subsection{Training}
Falcon was trained on the ShareGPT dataset, utilizing 68,000 dialogue iterations. We employed the AdamW optimizer with beta values ($\beta_1$, $\beta_2$) set to ($0.9$, $0.95$) with a learning rate set to 3e-5. The settings are consistent with Eagle \cite{li2024eaglespeculativesamplingrequires}. The semi-autoregressive head is trainable within two days on an H800 server.

\begin{table}[ht]

\centering
\setlength{\tabcolsep}{1mm} 
\begin{tabular}{ccccc}
\toprule
 &  & MT-Bench & HumEval & GSM8K\\
\midrule
 Model& Method&speedup  &speedup&speedup\\
\midrule
\multirow{6}{*}{V7B}
 & SpS&1.82x&1.99x&1.71x\\
 &PLD&1.61x&1.82x&1.82x\\
 &Lookahead&1.63x&1.72x&1.84x\\
 &Medusa&2.06x&2.41x&2.22x \\
 &Eagle&2.82x&2.95x&2.72x\\
 &\textbf{Falcon}&\textbf{3.10x}&\textbf{3.21x}&\textbf{2.92x}\\
 \midrule
\multirow{6}{*}{V13B}
&SpS&1.93x&2.23x&1.77x\\
&PLD&1.58x&1.85x&1.68x\\
&Lookahead&1.65x&1.71x&1.81x\\
&Medusa&2.32x&2.44x&2.32x \\
 &Eagle&2.85x&2.97x&2.52x\\
 &\textbf{Falcon}&\textbf{2.97x}&\textbf{3.12x}&\textbf{3.13x}\\
 \midrule
 \multirow{4}{*}{LC7B} 
 &PLD&1.38x&1.52x&1.32x\\
 &Lookahead&1.61x&1.72x&1.58x\\
 &Eagle&2.71x&2.93x&2.89x\\
 &\textbf{Falcon}&\textbf{2.94x}&\textbf{2.95x}&\textbf{2.91x}\\
 \midrule
 \multirow{4}{*}{LC13B} &PLD&1.42x&1.63x&1.41x\\
 &Lookahead&1.61x&1.72x&1.58x\\
 &Eagle&2.95x&3.25x&2.93x\\
 &\textbf{Falcon}&\textbf{3.11x}&\textbf{3.51x}&\textbf{3.10x}\\
\bottomrule
\end{tabular}
\caption{Comparison among Falcon and other speculative decoding methods in terms of speedup ratio.}
\label{table_speed}
\end{table}

\begin{table}[ht]

\centering
\begin{tabular}{ccccc}
\toprule
 &  & MT-Bench & HumEval & GSM8K\\
\midrule
 Model& Method&$\alpha$  &$\alpha$  &$\alpha$\\
\midrule
\multirow{3}{*}{V7B}
 &Medusa&60.53&63.26&61.09 \\
 &Eagle&74.59&76.18&75.47\\
 &\textbf{Falcon}&\textbf{77.64}&\textbf{79.92}&\textbf{78.03}\\
 \midrule
\multirow{3}{*}{V13B}
&Medusa&61.77&64.58&62.39 \\
 &Eagle&74.91&76.26&75.46\\
 &\textbf{Falcon}&\textbf{77.60}&\textbf{80.34}&\textbf{80.22}\\
 \midrule
 
 \multirow{2}{*}{LC7B} 
 &Eagle&72.98&74.76&73.89\\
 &\textbf{Falcon}&\textbf{74.31}&\textbf{77.07}&\textbf{75.51}\\
 \midrule
 \multirow{2}{*}{LC13B}&Eagle&73.64&76.51&75.28\\
 &\textbf{Falcon}&\textbf{75.27}&\textbf{78.52}&\textbf{77.13}\\
\bottomrule
\end{tabular}
\caption{Comparison among Falcon and other speculative decoding methods in terms of Acceptance rate ($\alpha$).}
\label{table_alpha}
\end{table}

\subsection{Effectiveness}
Figure \ref{compare_figure} and Table \ref{table_speed} show the speedup comparison among Falcon and other state-of-the-art speculative decoding methods on MT-bench, HumanEval and GSM8K for greedy (temperature=0). Speedup ratios of Falcon, Eagle, and Medusa were fairly tested on an H800 server. The results of Speculative Sampling (SpS), Prompt lookup decoding (PLD), and Lookahead were copied from their technical reports. Falcon has achieved a speedup ratio ranging from 2.91x-3.51x. Compared to AR drafters, Lookahead and Eagle, Falcon is faster by  1.79x-2.06x and 1.01x-1.24x, respectively. This improvement is because of Falcon's SAR property; the draft model can generate multiple tokens each forward, which reduces the latency of each drafting phase and thus increases the total speedup ratio. Compared to Medusa, which is a SAR method as well, Falcon is faster by 1.24x-1.50x. It is attributed to the consideration of the dependency among tokens. Falcon adopts a CSGD method to maintain token dependency in the training stage. This enables the drafter to predict tokens more accurately than Medusa, effectively reducing the cost from the verification phase.

Table \ref{table_alpha} illustrates the Acceptance rate ($\alpha$) of Falcon, Medusa, and Eagle. The experiments were performed on the H800 server. Falcon outperforms Eagle by 3\% to 5\%, indicating that more tokens are generated by the draft model rather than the target LLM. Compared to the SAR method (Medusa), Falcon achieves a higher $\alpha$ by 15.76\%-17.83\%, which illustrates the importance of the token dependency in the draft phase.

Table \ref{table_tau} compares the average acceptance length ($\tau$). Although the Vicuna 7B model has a lower $\tau$ than Eagle, other sizes of models have shown a higher acceptance rate. It is demonstrated that the drafter of Falcon has a high prediction accuracy, even surpassing the AR models, and thus illustrates the effectiveness of CSGD. Compared to Medusa, the intrinsic design restricts it to conduct at most one forward each draft phase, which leads to a low $\tau$. However, there is no such limitation to Falcon. In the experiment, Falcon performs four forward passes each drafting phase, achieving a higher $\tau$ than Medusa by 1.78-2.22.

\begin{table}[h]

\centering
\begin{tabular}{ccccc}
\toprule
 &  & MT-Bench &HumEval & GSM8K\\
\midrule
 Model& Method&$\tau$  &$\tau$&$\tau$\\
\midrule
\multirow{3}{*}{V7B}&Medusa&1.51&1.71&1.55 \\
&Eagle&3.94&4.29&4.00\\
&Falcon&3.34&3.88&3.70\\
\midrule
\multirow{3}{*}{V13B}&Medusa&1.59 &1.81&1.64 \\
  &Eagle&2.91&3.13&2.97\\
  &\textbf{Falcon}&\textbf{3.37}&\textbf{3.97}&\textbf{3.86}\\
 \midrule
 
 \multirow{2}{*}{LC7B} 
 &Eagle&2.43&2.91&2.78\\
 &\textbf{Falcon}&\textbf{2.82}&\textbf{3.30}&\textbf{3.02}\\
 \midrule
 \multirow{2}{*}{LC13B}
 &Eagle&2.73&3.21&2.98\\
 &\textbf{Falcon}&\textbf{2.95}&\textbf{3.60}&\textbf{3.47}\\
\bottomrule
\end{tabular}
\caption{Comparison among Falcon, Eagle, and Medusa in terms of Average acceptance length ($\tau$).} 
\label{table_tau}
\end{table}                
\subsection{Ablation Study}
\subsubsection{Tree Attention}
Falcon, similar to Eagle and Medusa, employs tree attention in both drafting and verification phases, while methods like speculative decoding do not use it. To remove the effect of tree attention, we applied a chain to Falcon, whose length is equal to the tree height, in order to maintain the same forward times of the draft model. Table \ref{ablation study} shows the comparative results indicating the impact of tree attention. In Falcon, the implementation of tree attention mechanisms has been demonstrated to enhance the acceleration ratio by 1.12x, increase $\alpha$ by 1.22, and improve $\tau$ by 4.96\%. This improvement is attributed to the fact that the tree attention increases the number of tokens validated and accepted, thereby augmenting the overall token throughput.
\subsubsection{Coupled Sequential Glancing Distillation}
The SAR approach of Falcon enhances the token numbers of each forward at the cost of reduced accuracy. The CSGD training method was employed to mitigate the drop in precision. We conducted tests on Vicuna 7B under two conditions: training the SAR head with Eagle's token-shift method \cite{li2024eaglespeculativesamplingrequires} alone and with the CSGD method. Table \ref{ablation study} illustrates the impact of CSGD. The implementation of CSGD resulted in a 1.17x increase in acceleration ratio, 0.56 in $\tau$, and 3.26\% improvement in $\alpha$. The above results demonstrate that CSGD can significantly improve the drafter's accuracy.
\begin{table}[h]

\centering
\begin{tabular}{cccccc}
\toprule
 Model&TA & CSGD & speedup &$\tau$ & $\alpha$\\
\midrule
\multirow{4}{*}{V7B}& $\times$ &$\times$ &2.06x&1.64&62.86\\

&$\times$ & $\checkmark$ &2.27x&2.13&68.68\\
&$\checkmark$ &$\times$ &2.65x&2.78&74.37\\
&$\checkmark$ & $\checkmark$ & \textbf{3.10x}  & \textbf{3.34}&\textbf{77.64}\\
\bottomrule
\end{tabular}
\caption{The ablation study of Falcon of the impact of Tree Attention (TA) and CSGD on MT-Bench.} 
\label{ablation study}
\end{table}

\subsubsection{$k$ factor}
The $k$ factor determines the number of tokens the drafter generates in a single forward pass, which is important to SAR drafting. We tested conditions with $k=\{2,3,4\}$ on Vicuna-7B. Considering the accuracy descending of SAR, we adopted fewer forward pass times in the drafting phase for $k=\{3, 4\}$ to achieve speedup ratios as high as possible. The corresponding heights of the decoding trees are 6 and 8, respectively. The results are presented in Table \ref{different k factors}. With the increasing of the $k$ factor, we see a drop in all three metrics, but the speedup ratios are still higher than Eagle (2.82x) and Medusa (2.06x). This is due to the reduced time it takes for an SAR drafter to pass forward with the same number of tokens generated by AR drafting, indicating the inherent advantages of enhanced SAR drafting. 
\begin{table}[h]

\centering

\begin{tabular}{ccccc}
\toprule
 Model&$k$ & speedup &$\tau$ & $\alpha$\\
\midrule
\multirow{3}{*}{V7B}& \textbf{2} & \textbf{3.10x} & \textbf{3.34}  & \textbf{77.64}\\
&3 &2.96x&2.52&72.21\\
&4 &2.86x&2.35&70.78\\
\bottomrule
\end{tabular}
\label{different k factors}
\caption{The comparison of $k$ factors on MT-Bench.} 
\end{table}
\section{Conclusion and Future Work}

In this paper, we propose Falcon, which utilizes the Coupled Sequential Glancing Distillation method to bolster the inter-dependencies among tokens, thereby enhancing the quality of the drafts. We also provide a theoretical analysis that clarifies the inner workings of our method. Additionally, we have developed a specialized decoding tree to support SAR drafting, which increases the token acceptance rate. Comprehensive evaluations indicate Falcon's superior performance over the Vicuna and LLaMA2 model series. On the MT-bench, HumanEval, and GSM8K datasets, Falcon is 2.91x-3.51x faster than vanilla Transformer, while maintaining a compact size comparable to two Transformer layers. Falcon appeals to applications that demand real-time responses and have limited computing resources.

The main challenge in accelerating the LLMs through speculative decoding lies in improving the accuracy and efficiency of the drafter under resource-limited conditions. Therefore, our future research efforts will focus on developing drafters that attain a high token acceptance rate with minimal overhead. This goal will be met by advanced semi-autoregressive or potentially non-autoregressive decoding techniques that fortify the interdependence among tokens. In addition, the dynamic modification of the decoding tree is another avenue that warrants further exploration.

\bibliography{falcon}

\appendix

\section{Tree Structure}

\begin{figure*}[htb]
    \centering
    \includegraphics[width=2\columnwidth]{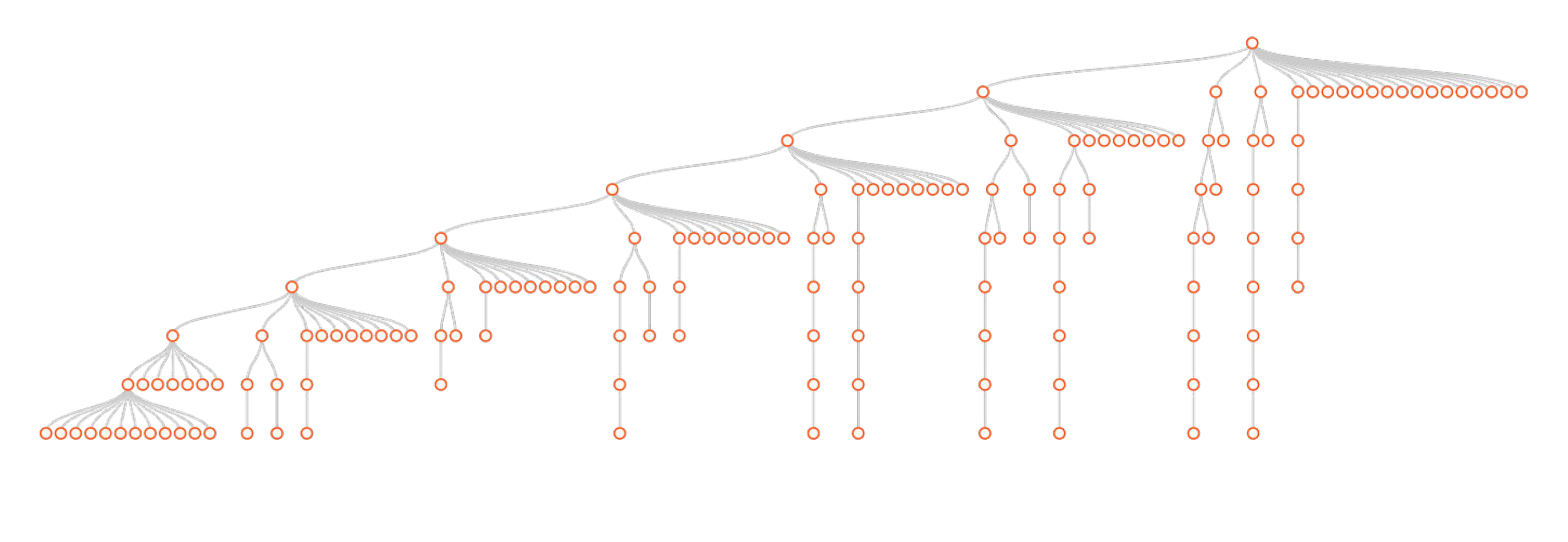}
    \caption{The tree structure used for Vicuna series.}
    \label{fig:vicuna tree}
\end{figure*}

\begin{figure*}[htb]
    \centering
    \includegraphics[width=1.85\columnwidth]{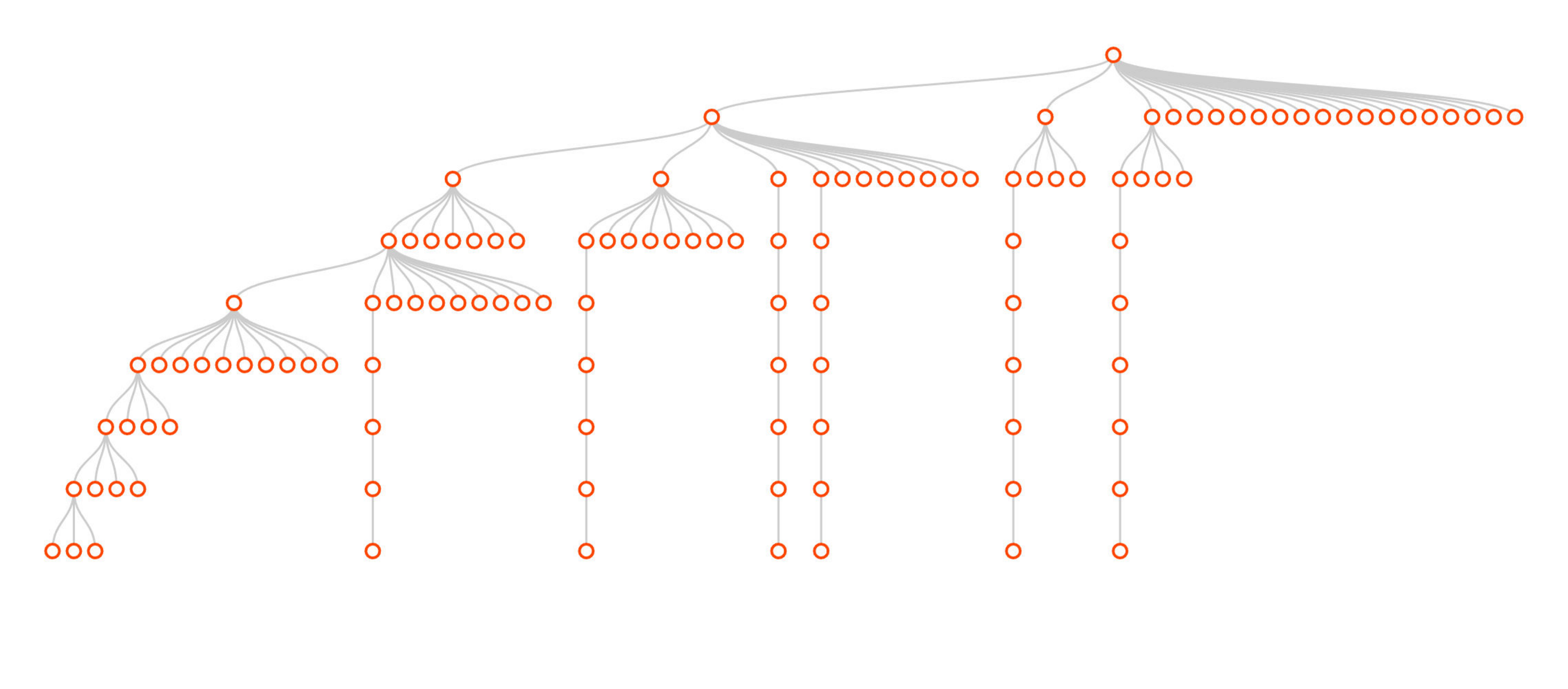}
    \caption{The tree structure used for LLaMA series.}
    \label{fig:llama tree}
\end{figure*}

A custom-designed decoding tree is introduced to accommodate SAR drafting. Figure \ref{fig:vicuna tree} and figure \ref{fig:llama tree} espectively display the tree structures utilized by the Vicuna models and the LLaMA models. It's important to note that, apart from the root node, each node on the tree represents only one token. The number of tokens for the root node is determined by k (the number of tokens initially generated by LLM or produced by the draft model in each forward pass). In a greedy setting, we select the top $m$ tokens with the highest probabilities as child nodes. In a non-greedy setting, we sample m tokens. The number of child nodes, m, can be inferred from Figure 5;
for instance, $m = 20$ at the root node. For different nodes, the value of $m$ can vary.  For trees of the same depth and $k$, the draft model undergoes the same number of forward passes during the draft phase. For example, whether it's Figure \ref{fig:vicuna tree} or Figure \ref{fig:llama tree} of these trees, when $k$ equals 2, the draft model performs 4 forward passes.

\textbf{Why do we use such a complex tree structure?} The choices of tree structure are optimized based on intuition and findings during experiments: branches of higher-probability tokens should be deeper and wider. We also find that the optimal tree structure is likely both
context-dependent and model-dependent. For instance, the tree structure that is effective for the Vicuna models may not necessarily yield the best acceleration results when applied to the LLaMA models, thus we design different tree structures for these models.

\end{document}